\documentclass[letterpaper, 11pt]{article}

\usepackage[english]{babel}
\usepackage[utf8x]{inputenc}
\usepackage[T1]{fontenc}

\usepackage[letterpaper,top=1in,bottom=1in,left=1in,right=1in,marginparwidth=1.75cm]{geometry}

\usepackage[numbers,sort]{natbib}
\usepackage{hyperref}
\usepackage{amsmath}
\usepackage{graphicx}
\usepackage[colorinlistoftodos]{todonotes}
\usepackage{soul} 
\usepackage[colorinlistoftodos]{todonotes}
\usepackage[linewidth=1pt]{mdframed}
\usepackage{wrapfig}
\usepackage{subcaption}

\newcommand{\G}{$\mathcal{G}$}
\newcommand{\Gp}{$\mathcal{G}^+$}
\newcommand{\nicebox}[1]{
  \begin{center}
    \begin{mdframed}[backgroundcolor=black!10]
        \noindent #1
    \end{mdframed}
    \noindent
  \end{center}
}

\title{\vspace{-65px}How Should AI Interpret Rules? A Defense of Minimally Defeasible Interpretive Argumentation} 
\author{John Licato\\
\texttt{licato@usf.edu}\\
Advancing Machine and Human Reasoning (AMHR) Lab\\
Department of Computer Science and Engineering\\
University of South Florida\\\\
\textbf{Draft version, last updated:} \today\\
Extended version of a talk given at USF AI+X Seminar, Oct. 29, 2021}

\date{}
\begin{document}
\maketitle

\vspace{-5px}

\noindent Can artificially intelligent systems follow rules? The answer might seem an obvious `yes', in the sense that all (current) AI strictly acts in accordance with programming code constructed from highly formalized and well-defined rulesets. But here I refer to the kinds of rules expressed in human language that are the basis of laws, regulations, codes of conduct, ethical guidelines, and so on. The ability to follow such rules, and to reason about them, is not nearly as clear-cut as it seems on first analysis. Real-world rules are unavoidably rife with \textit{open-textured terms}, which imbue rules with a possibly infinite set of possible interpretations. Narrowing down this set requires a complex reasoning process that is not yet within the scope of contemporary AI.

This poses a serious problem for autonomous AI: If one cannot reason about open-textured terms, then one cannot reason about (or in accordance with) real-world rules. And if one cannot reason about real-world rules, then one cannot: follow human laws, comply with regulations, act in accordance with written agreements, or even obey mission-specific commands that are anything more than trivial. But before tackling these problems, we must first answer a more fundamental question: Given an open-textured rule, what is its \textit{correct} interpretation? Or more precisely: How should our artificially intelligent systems determine which interpretation to consider correct? In this essay, I defend the following answer: \textbf{Rule-following AI should act in accordance with the interpretation best supported by \textit{minimally defeasible interpretive arguments} (MDIA).} In Section \ref{sec:interpretive}, I will introduce and defend the need for AI which can produce interpretations of rules that are supported by \textit{interpretive argumentation}, and advocate for AI that is \textit{argument-justified} rather than merely \textit{explainable}. But in order to recognize interpretive argument quality, we need a way to estimate argument strength. Section \ref{sec:min_defeasibility} argues that the \textit{minimal defeasibility} standard is our best hope. Finally, I address anticipated counterarguments and possible points of confusion in Section \ref{sec:objections}.

\section{Introduction: Why Addressing Open-Texturedness Matters}
\label{sec:intro}

An autonomous guard robot, \G, stands guard at a small recreational park in a quiet suburban neighborhood. \G\ was purchased by the local homeowner's association (HOA), from a new tech startup that makes general-purpose guard robots for broad use. Configuring these robots is ostensibly no more difficult than opening up an app, typing in the rules or ordinances the robot must enforce, and then pressing `Start.' The HOA representatives do so, tasking \G\ with one extremely simple rule: ``No vehicles in the park.''

This rule, according to the HOA representatives, seems straightforward enough to serve as a initial test case. But the very next day, \G\ spots a child running with a plastic toy car around just inside of the park's main entrance. \G\ immediately snatches the toy, and throws it into the nearest vehicle-allowing area outside of the park. The child's parents are outraged, so the HOA dispatches the college-aged son of an HOA member, so chosen because of his previous successes with ``tech-related stuff.'' Dubbed ``tech guy,'' he immediately sets about re-working the rules uploaded to \G\ in two ways: It must specify that toy cars are allowed, and it must respond to violations of the rule appropriately in the future. ``Easy enough,'' says tech guy, who subsequently opens up the app and edits its sole rule to become: ``No motorized vehicles in the park. Violations will be rectified by \G\ in an appropriate manner.''

The next day, two very different cases come to \G's attention: a person in a motorized wheelchair enters the park, followed by a person in a full-sized, ear-shatteringly-loud motorcycle. But in both cases, \G\ does absolutely nothing. When tech guy checks \G's logs later that night to find out why, he learns that although \G\ decided that both the wheelchair and the motorcycle violated the rules (the former of these an obvious mistake that the HOA member decides to deal with later), \G\ did not know how to determine what was an ``appropriate manner'' of enforcement. After all, politely and quietly asking the motorcycle owner to leave would likely be ignored (unintentionally or intentionally), and forcefully evicting the wheelchair owner would be an application of force much like \G's previous solution to the child and the toy car---and that action, \G\ inferred, was inappropriate, although \G\ is hard-pressed to explain why. Tech guy is beginning to learn a lesson that will be an ongoing theme of his experience: \textit{the full background needed to properly interpret rules cannot be transmitted via the text of those rules alone.}

Tech guy now appeals to the company which made \G, who promptly responds by uploading a software patch using the latest advances in artificial intelligence, though they warn that the patch is highly experimental. It works, they explain, by using an extremely powerful statistical reasoning algorithm known as a ``deep neural network,'' which has been pre-trained on an input of hundreds of thousands of instances of similar park-related rules throughout the country being enforced. This training data was laboriously and manually labeled, so that every enforcement action described within was confirmed to be either appropriate or not, by a highly qualified but underpaid panel of first-year law students. Tech guy gladly installs the patch, and in order to prevent the wheelchair misclassification, edits \G's rules to: ``No motorized vehicles in the park unless they are necessary for mobility. Violations will be rectified by \G\ in an appropriate manner.''

Now \G\ seems to be doing much better. Its deep neural network patch has given \G\ an ability to determine the level of force necessary is some function of the severity of the violation and the immediate danger the violation poses, which in most cases amounts to not much more than a volume-modulated reminder that vehicles are not allowed in the park. The update even demonstrates an ability to correctly classify some items it hasn't seen before as vehicles and non-vehicles. For example, a biker rides a shiny new motorcycle into the park of a make and model nobody in town has ever seen before. \G\ correctly identifies it as a vehicle, thus seemingly displaying an unprecedented robustness. Tech guy is lauded as a hero by the HOA committee.

But one day, the HOA president calls tech guy in a panic. The latest motorcycle banning, she explains, completely misunderstands the rule. After all, the rule was instituted in the first place because the HOA had received complaints that all the gas-guzzling vehicles in the park were too loud, disturbing the rest of the residents. The recently-banned motorcycle, on the other hand, was fully electric, and much quieter. Thus, because \G\ did not understand the \textit{historical reasons} for the rule, and did not have any insight into the \textit{intentions} of the HOA members who instituted the rule, it was not able to properly interpret it. 

But even worse, the HOA president continues, today someone was severely injured in the park. When an ambulance attempted to enter the park to get to that person, \G\ stood in its way. When the ambulance techs forcefully pushed past \G, it interpreted this as an escalation of force, and pushed back! The ambulance was able to reach the injured person only after fooling \G\ into believing there was a person with a tank entering on the other side of the park. Clearly, \G\ did not have a proper understanding of what might constitute a reasonable exception to the vehicles rule. 

Furthermore, \G\ seems to be inconsistent. A gang of mini-moped riders entered the park the other day. Half of them were asked to leave, the other half were ignored. Tech guy immediately dives into \G's logs and discovers that for each mini-moped, \G\ classified each as a vehicle with confidence values near 50\%---and because some barely made the cut (50.01\%), those were banned, whereas the others (49.99\%) were not. The reason for this variation was something that the tech-competent member could \textit{explain but not justify}; it may have had something to do with minor differences in each mini-moped's colors, noise level, rider appearance, etc. 

To rectify this, the HOA propose a new update to the rule: \textbf{``All motorized devices which produce sounds of over 75 dB are not allowed in the park. Reasonable exceptions for ethical or emergency reasons will be allowed at the discretion of \G."} Tech guy is tasked with inputting this new rule into \G's app, and ensuring that it ``enforces the rule with no problems,'' the HOA president confidently commands. But by now, poor tech guy knew this was no simple task: if \G\ had an issue properly interpreting the term `vehicle,' how well is it going to do with `device,' `produce sounds,' and most importantly, ``reasonable exceptions''? He reaches out to \G's manufacturer, who then informs him that the deep neural network used to interpret the term `vehicle' is no longer of use here---entirely new datasets will need to be collected and new deep neural networks re-trained and tested. 

\paragraph{The Inevitability of Open-Texturedness} Let's take a step back now and try to extrapolate lessons learned. Students of law may recognize the vehicles in the park example on which tech guy's story is based. It was made famous by influential legal philosopher H.L.A. Hart \cite{Hart1961}, and is often invoked to illustrate the existence of \textit{open-textured terms} in rules. Open-texturedness \cite{Waismann1965} refers to ``the fact that however tightly we think we define an expression, there always remains a set of (possibly remote) possibilities under which there would be no right answer to the question of whether it applies'' \cite{Blackburn2016}. We'll refer to rules as \textit{open-textured rules} if they contain, and their correct interpretations depend on, open-textured terms. In our running example, the term `vehicle' is open-textured---although there are many instances of things that are clear vehicles (a 2001 Volkswagen Jetta) or clear non-vehicles (a bolt of lightning), there are always ways to think of objects that are borderline cases. This is the case no matter how well you happen to define them, as those definitions will themselves contain open-textured terms. And even if you produce a complex definition that seems to cover all conceivable cases that one might realistically ever encounter, many other problems may arise: the definition may not exactly align with the intentions of the rule's creators, or the historical context of the rule, or might rely on outdated concepts, and so on \textit{ad infinitum}. 

In fact, we can go so far as to say that open-texturedness is a necessary, unavoidable, and \textit{desirable} feature of rule systems. How can this be true? Doesn't the park example above highlight shortcomings with open-textured rules that we are best off avoiding at all costs? Certainly, awareness of open-texturedness may lead us to create better-worded rules to minimize their negative effects (this is why Hart's park example is taught to all students of law). But open-texturedness can not be avoided in real-world rules. As USSC Justice Oliver Wendell Holmes once cautioned, ``the machinery of government would not work if it were not allowed a little play in its joints'' \cite{Holmes1931}. Lawmakers often intentionally use vague language, in part because it allows for the delegation of discretion to boots-on-the-ground agents \cite{Staton2008,Asgeirsson2020}.\footnote{Vagueness is not quite the same as open-texturedness, but the benefits we state here overlap.} Insofar as they provide the means for flexibility in interpretation, open-textured terms are an unavoidable and necessary feature of legal, ethical, and policy regulations \cite{Hart1961,Franklin2012,Prakken2017,Quandt2020,Licato2019b,Licato2018c}. 
As such, autonomous reasoning software systems at all levels must learn how to work with them.

Well then, perhaps terms dealing with moral or otherwise abstract concepts are unavoidably open-textured. Can we sidestep open-texturedness by only allowing AI to reason over rules that do not contain such predicates? Unfortunately, the problem of open-texturedness persists even in domains with seemingly concrete or banal terms. For example, consider a traffic regulation stating that vehicles must ``keep to the right as far as is \textit{reasonably safe}'' \cite{Prakken2017}. Such a regulation would require interpretation by autonomous driving vehicles. But it is implausible to exhaustively list an exception-free accounting of all possible scenarios and conditions that can be considered instances of the open-textured term `reasonably safe'---any such attempt would inevitably limit the scope of the regulation and render it fatally inflexible in the face of unanticipated conditions. One might note that in practice, such banal terms may be open-textured, but are rarely a point of dispute in courts. The commonsense notion of `reasonably safe' as it appears in this context is similar enough amongst most people that it is clear to many what does and does not fall under its description. But this is the point to be highlighted exactly: the ability to reach this `commonsense' understanding is not something that current state-of-the-art AI can do reliably, and we seek an approach that can guide AI research in that direction.

In short: rules that are too formal and specific tend to be difficult to understand / communicate, and tend to have such a narrow domain of applicability as to make them of little use in complex domains. It is likely impossible that rules in any representational system, when those rules must govern behavior in complex real-world domains, and the rules must have a reasonable degree of human-understandability, can completely avoid open-textured concepts (and complete formal rigidity in such rules may not be preferable anyway \cite{Guarini2006,Anderson2007b,Quandt2019,Licato2018c}).
Consider the language used in rules in ethical and legal domains. Whether they work fully autonomously or in human-machine teams, artificial agents given rules to follow (for rules ranging from international laws, to company ethical policies, to mission-specific orders) can benefit tremendously by understanding how to use interpretive reasoning to determine the applicability of open-textured phrases \cite{Quandt2019}. For example, the ACM/IEE-CS Software Engineering Code of Ethics \cite{Gotterbarn1997} states that software engineers should ``[m]oderate the interests of the software engineer, the employer, the client and the users with the public good.'' But the phrase `public good' is highly open-textured, and people may disagree about whether certain plausible actions are in service of the public good. If a software engineer creates software that destroys all of the world's computers, should that be considered in service of the public good? Indeed, the engineer might offer interpretive argument $I_1:$ ``Destroying all computers allows humanity to return to a state of nature, which is a good thing.'' Such an argument would be quickly dismissed by most human reasoners. But on what basis is such a dismissal warranted? And more importantly for artificial intelligence researchers: how can interpretive arguments like $I_1$ be automatically evaluated in a human-like way?

\paragraph{Benefits of interpretation-capable AI} We can phrase the problem of open-texturedness in AI as follows: \textit{How can we ensure rule-following AI interprets open-textured rules correctly?} And immediately we see a problem: the question itself contains the highly open-textured term `correctly.' What does it mean for an interpretation of an open-textured rule to be correct? Our answer to this question will be that `correctness' in this question can be approximated by finding the interpretation that is supported by minimally defeasible interpretive arguments. 

But before we introduce those concepts, let us reflect a bit on what kinds of technologies might emerge from the kind of AI that we have in mind. If an artificially intelligent reasoner is capable of determining whether an interpretation of a set of open-textured rules is correct (according to some normative standard of correctness in interpretation), then we will refer to it as an \textit{interpretation-capable AI}. If we advance the state of AI so that it is interpretation-capable, there are at least three categories of significant advances.

\ul{First}, it can lead to tools to help the development of rules. Regulatory agencies, from complex government agencies to the HOA in our above example, can write rules and have those rules examined by interpretation-capable AI. Possible ambiguities, or other loopholes in the rules, can be discovered by the AI and the writers can improve the rules accordingly. For example, in the domain of cybersecurity, access policies may consist of open-textured rules defined by decision-making committees, which must then be translated into computer-readable code by programmers. Ensuring this translation was performed correctly could be carried out with the aid of an interpretation-capable AI.

\ul{Second}, it can lead to tools to help the enforcement and understanding of rules. When rulemakers write their rules, they have intentions and historical context that may not come across in the text of the rules themselves. But the people who must follow or enforce the rules (such as \G\ and the residents of the park's commmunity) may only have easy access to the text of the rules. Think about the laws and regulations that you must follow at work and in your community---you may broadly know what many of them permit and do not permit, but do you have a full understanding of \textit{why}? Or, as is more likely, are there some rules that you simply follow without knowing the reasoning behind them? Thus, this misalignment between how the public might interpret the text of some rule and how it is meant to be interpreted might be minimized with the assistance of an interpretation-capable AI.

But the understandability of rules can be advanced in another way as well. In theory, every action you perform is potentially subject to thousands of rules, laws, and regulations at the levels of local, state, national, international, trans-galactic, etc. In practice, we are often safe by just paying attention to rules in domains that seem only superficially relevant, but this strategy is not fool-proof---for example, how was Al Capone to know that he would ultimately be found guilty of violating tax laws? Indeed, there are some domains (tax law among them) where the set of rules that might plausibly govern any given action is so large that teams of dedicated professionals are needed to determine legality, and even then they miss things. 

\ul{Finally}, and perhaps most consequentially, interpretation-capable AI may further the \textit{democratization of legal protection}. Textual interpretation is not a trivial reasoning task, one made significantly more difficult by facts such as: only interpretations made in accordance with norms of legal practice, and there is an almost innumerable quantity of existing laws and regulations governing us all. It is no surprise that competent legal practitioners and experts are in high demand (and can command salaries to match). But this means that access to high-quality legal advice, despite the best efforts of our current judicial systems, can be uneven for people with different socioeconomic statuses. The wide availability of interpretation-capable AI may reduce this inequity.


\section{Interpretive Argumentation}
\label{sec:interpretive}
At this point, I've hopefully convinced you that interpretation-capable AI is both desirable and necessary (at least if we ever want AI to follow the rules we create for them). But how can we create interpretation-capable AI? 

Let's return to the example of \G. Resigned to the absurdity of his Sisyphean task, tech guy returns to the park. He dutifully inputs the new rules into \G, as instructed by the HOA president, and goes home for the day. But now a group known as ``Scooters for Charity'' enters the park with gas-powered motorized scooters, hoping to have their annual fundraiser picnic. \G\ approaches them to measure their noise levels, but the leader of the group explains that even if their scooters are too loud, they must be considered a reasonable exception: Each scooter emits sounds not above 75.1 dB, and the picnic is for a good cause. Poor \G\ is stumped again.

Instead of going into another iteration of guess-and-refine, let's ask ourselves: how would we want \G\ to consider this exception request? A reasonable way forward is to ask \G\ to consider arguments for and against, a list which might (in the ideal case) look something like this:

\begin{mdframed}
\footnotesize
  \textbf{Arguments in favor of an exception:} 
     \begin{itemize}
        \item \textit{Arguments from ordinary meaning}: The purpose of the picnic is for charity, and the common understanding of ``ethical reasons'' encompasses activities that support charities.
        \item \textit{Arguments from precedent}: In the past, extremely minor violations of noise ordinances were allowed if the event was only held annually, as this picnic is.
        \item \textit{Arguments from analogy}: On a road with a 75 MPH speed limit, it is considered reasonable to not pull over drivers going 75.1 MPH. Therefore, 75.1 dB should be considered close enough to 75 dB to warrant an exception.
        \item \textit{Arguments from statutory history}: Historically, this group's charity picnics have been allowed in the park and were considered by the HOA to be ethical.
        \item \textit{Arguments from statutory purpose}: The reason for the law was to protect the enjoyment of the park for residents, and the picnic is an enjoyable activity for the residents.
        \item \textit{Arguments from intended meaning}: The authors of the law did not intend the rule to stifle charitable activities.
    \end{itemize}
\end{mdframed}

\begin{mdframed}
\footnotesize
  \textbf{Arguments against an exception:} 
     \begin{itemize}
        \item \textit{Arguments from ordinary meaning}: Motorized scooters are plainly ``motorized vehicles.''
        \item \textit{Arguments from technical meaning}: A noise output of 75.1 dB is technically above the threshold of 75 dB.
        \item \textit{Arguments from legal principle}: \G\ has a duty to be consistent in its interpretations and enforcement of the rules, and an exception will endanger that.
        \item \textit{Arguments from statutory purpose}: The reason for the rule was specifically to limit noise levels, and an exception in this case would violate that purpose.
        \item \textit{Arguments from intended meaning}: The exception clause in the rule was intended to be invoked extremely rarely, and for extremely urgent reasons only.
    \end{itemize}
\end{mdframed}


Of course, this list is far from complete. And stronger versions of each of these arguments might be formulated, given enough time and resources. But we might imagine that if \G\ were an impartial agent hoping to produce a thorough, explainable, justifiable interpretation of its rules, it would be able to produce and evaluate argument lists like the above. You will notice that the arguments tend to fall rather nicely into categories, some of which are explicitly named. This is no coincidence: arguments that can be made for or against interpretations of a textual rule are called \textit{interpretive arguments}. MacCormick et al. (1991) \cite{MacCormick1991} identified eleven types of interpretive arguments commonly used in legal argumentation, and their categorization has since been built on and refined in various ways \cite{Rotolo2015,Loui2016,Walton2016b,Pereira2017,Macagno2018b,Walton2018}, some of which are reflected in our above lists. In general, interpretive arguments, following the general formulation of Sartor et al.\ \cite{Sartor2014} are of the form: ``If expression $E$ occurs in document \textbf{D}, $E$ has a setting of $S$, and $E$ would fit this setting of $S$ by having interpretation $\mathcal{I}$, then $E$ ought to be interpreted as $\mathcal{I}$'' \cite{Sartor2014}. Thus understood, `interpretive arguments' are those used to support or attack an interpretation of a fixed expression within a fixed document. 

Note that the categories of interpretive arguments allow for more nuanced examinations into why an interpretation should or should not be accepted. For example, an interpretation is never considered correct merely because it is similar to previous cases. Instead, the \textit{relevance of that similarity} must be made explicit. Does the similarity have the force of an analogical argument? Does it set a binding precedent? Does it establish a commonsense understanding of a word or term? Each of these ways of translating similarity into argumentative force are associated with a specific category of interpretive argument.

Now, generating such a list of interpretive arguments is already something that current state-of-the-art AI is unable to do consistently. Progress is certainly being made in this area---current work involves a combination of argument mining and template-based generation (e.g., \cite{Walton2012,Habernal2014,Wachsmuth2016,Lippi2016,Lawrence2016,Hou2016,Reed2017,Moens2017,Habernal2017}). But even if AI can generate an approximately complete list of interpretive arguments, at least two more complex problems arise: (1) How should \G\ evaluate the strength of these individual arguments? And (2) how should \G\ combine them to produce a final judgment? These are questions of central importance to argumentation in general, and any potential answer to them would likely not be satisfying to all people in all cases. But when human judges, law-enforcement agents, or everyday rule-followers are faced with such decisions, they often cannot afford to spend their entire lives carrying out intensive philosophical research. They must weigh the arguments, \textit{somehow}, and formulate a final decision. And so must interpretation-capable agents such as \G.

\textbf{Argument-Justified AI vs.\ Explainable AI} We will discuss a way to evaluate and combine such arguments in Section \ref{sec:min_defeasibility}. Let us assume for now that we have \textit{some} satisfactory method of doing so, and that tech guy programs \G\ with this ability, thus yielding the ultimate textual interpretive reasoner \Gp. We will have then, ostensibly, a way to resolve open-textured terms in rules: Given possible interpretations of that open-textured term, enumerate the strongest interpretive arguments for and against each interpretation, and then combine them. Then, select the interpretation that has the most favorable combination of overall interpretive arguments. Ignoring for now the substantial technical challenges in creating such an AI, \Gp\ would (by definition) be interpretation-capable. But more than that---such a system would produce interpretations that are \textit{argument-justified}, meaning that the interpretations it considers and settles on will be accompanied by: (1) a set of arguments for why the final interpretation should be considered correct, and (2) responses to counter-arguments against the final interpretation. Thus, I argue:
\nicebox{\textbf{CLAIM 1:} Rule-following AI should be argument-justified, striving to find the interpretation that it can support using the strongest possible interpretive arguments.}
Let us consider the merits and demerits of argument-justified AI as opposed to its alternatives. \textit{Explainable AI} (XAI) comes first to mind, as it is an active area of research in machine learning. XAI work takes the outputs of black-box systems and produces explanations for them. Although there are some overlaps between explanations and arguments, and the two can productively be used in combination with each other \cite{Bex2016}, there is a fundamental difference: \textit{explanations help people understand how an output was generated, while arguments persuade people that an output should be accepted.} 

Let us assume that in the future, someone comes up with a purely statistical deep neural network where all we have to do is feed as inputs: the rules to be followed, a description of the scenario to be interpreted, and some minimal set of contextual details so that the system can infer things like intents of the rule-makers, historical interpretations of the rules, etc. This system is an almost impermeable black box, and its outputs are somewhat explainable, but they are not argument-justified. Instead, this system (let's call it $\mathcal{O}$ for `oracle') simply outputs the optimal interpretation same as \Gp; i.e., the interpretation that would have come about if the best possible interpretive arguments of all types were generated and combined in an optimal way. $\mathcal{O}$ may even output a percentage that might be understood as a measure of confidence. Let us assume, for the sake of simplicity, that if $\mathcal{O}$ outputs an interpretation and a confidence of 50\% or higher, then the interpretation is ``recommended.'' Now ask yourself: If $\mathcal{O}$ were to exist today, and it produced the same interpretations as an argument-justified, interpretation-capable system, would it be preferable? 

I argue that $\mathcal{O}$ would \textit{not} be preferable to an equivalent argument-justified, interpretation-capable system. I would not, however, argue that the creation of $\mathcal{O}$ is impossible. It is conceivable that in the future a massive, well-designed artificial neural network could exactly simulate the brains of the 15 greatest Supreme Court justices who ever lived, simulate a lengthy and productive debate between them, and then run iteratively until a conclusion is reached. Presumably, such a system (or another similar brute force approach) would come as close as any other decision-making algorithm to coming up with the ``correct'' interpretation in the largest number of cases. But what I do doubt is that any approach to designing $\mathcal{O}$ can do so without, at some stage of its deliberations, internally generating and evaluating interpretive arguments. If $\mathcal{O}$ were able to generate an interpretation without carrying out any of these steps, then in all likelihood, it has failed to consider some crucial argument or counter-argument, and is therefore suboptimal as compared to \Gp\ (in the sense that it does not come up with the most correct interpretations). On the other hand, if $\mathcal{O}$ internally generates and evaluates interpretive arguments just like \Gp, then it is difficult to see why it should not simply provide the optimal interpretive arguments, along with the reasoning behind their combination it evaluated internally---but this makes it argument-justified anyway.

Now even if I am wrong about my claims in the previous paragraph, $\mathcal{O}$ would still not be preferable to \Gp, for several reasons. First, interpretations of open-textured rules must be subject to stakeholder analysis and approval. The interpretive argument paradigm provides a rich tapestry of justification types, and it is easy to see why interpretations that are justified with clearly laid-out interpretive arguments is preferable to a simple black-box output. Even if $\mathcal{O}$ were the most powerful pattern recognizer in existence, trained on the largest data set possible, if $\mathcal{O}$ is unable to argue \textit{why} we should accept its outputs, it will fail to persuade stakeholders. 
Further, there is a sense in which the correct answer to certain interpretive scenarios does not even exist until the stakeholders consider arguments for an interpretation. For example, the United States Supreme Court is not a legislative body, but when they decide on an interpretation of some open-textured term in a law, that interpretation is binding upon lower courts and also the Supreme Court itself, according to the principle of \textit{stare decisis}. Therefore, when interpreting law, is the Supreme Court merely \textit{discovering} correct interpretations that were always true, or are they \textit{creating} the correct interpretation through an interaction of values, viewpoints, and arguments? For our purposes, it will suffice to say that it is likely some combination of these two (I elaborate more on this idea in Section \ref{sec:objections}). 
And likewise, when \Gp\ encounters new scenarios that were not anticipated by its programmers, a complex medley of inference, value-laden judgments, and argument consideration interact, together shaping the interpretation that is ultimately accepted as the correct one. An interpretive framework which therefore fails to take any of these elements into account will be sub-optimal.

Another reason to prefer argument-justified AI relates to the even application of rules. That rules should be applied equally to all is a principle that is central to virtually all modern legal and ethical systems. If the same rule is assigned two different interpretations for two different target cases, the reasons for this difference must be made clear in such a way that they create a guide for future cases. In the example we saw earlier with \G\ and the motorized scooter group, suppose that \G\ was replaced with $\mathcal{O}$, and decided to grant an exception to the scooter group without providing substantial interpretive argumentation to support this decision (internally, the reason it decided to do so was because its internal statistical algorithm estimated a 50.01\% confidence that an exception was warranted). But then the next day, a different motorized scooter group arrives and decides to host an impromptu picnic, this time for a different charitable cause. Is $\mathcal{O}$ required to grant their request? If not, why not? And how can such questions be answered in the first place, in the absence of interpretive arguments? If the second group's request were to be denied (for example, perhaps $\mathcal{O}$'s internal algorithm only had a 49.99\% confidence that an exception was warranted), can we really say that $\mathcal{O}$'s judgements constitute a fair application of the rules across the two scooter groups?

\Gp, on the other hand, would be able to decide whether the second group should be awarded an exception \textit{on the basis of the network of interpretive arguments used to support its decision on the first group}. For example, it may be that the first group was granted an exception primarily because charitable events are good, ethical things to support. Thus, since the second group is also supporting charity, the exception should also apply. On the other hand, if the most influential reason for the first group's exception was that their proposed event was rare and the citizens of the park would not mind a single day's worth of noise, then rejecting the second group might be warranted. In this case, the second group might have cause to complain to the HOA, but at least when the HOA examines the case they will be able to consider the arguments \Gp considered. Either way, if \Gp\ or $\mathcal{O}$ are to apply laws equally and fairly across multiple circumstances, they must be able to demonstrate why interpretations across multiple borderline cases are consistent---and this can best be done with explicit rationales.

We can go on listing examples of what might happen in this park example: What if new information comes to light about a previous interpretation that forces revision? What if the socially-accepted meaning of one of the open-textured terms changes? What if a law is passed banning a certain type of justification for interpretations? What if bias is found in the dataset used to train $\mathcal{O}$? And so on, \textit{ad infinitum}. Real-world rules must contend with real-world scenarios, the complexity of which can quickly go beyond what was anticipated by the rule's creators. But this complexity is a key reason why open-textured rules are so prevalent in the first place: open-texturedness provides flexibility to interpretive reasoners. Considerations like this make it difficult not to conclude that any agent, artificial or natural, must be able to perform interpretive reasoning if they are expected to properly follow open-textured rules. 

\paragraph{Asimov's Laws Versus Argument-Justified AI} One more example before moving on. How do the ideas stated in this paper align with arguably the most famous example of interpretive reasoning? For those who are unfamiliar (spoiler alert!), one of the most famous creations of science fiction writer Isaac Asimov is known as his \textit{Three Laws of Robotics}, the first of which is often stated, ``A robot may not injure a human being or, through inaction, allow a human being to come to harm.'' What inevitably happens, as did in the Will Smith movie ``I, Robot'' (2004), is that the robots interpret this rule (due to the open-textured term `harm') as an imperative to take control of humanity, for humanity's own protection. 

What might an argument-justified version of the robots' interpretation look like? We might imagine that, if the individual modules used to generate each interpretive argument type were working properly, arguments for and against world conquest might look something like this:

\begin{mdframed}
\footnotesize
     \begin{itemize}
        \item \textit{Arguments from larger purpose}: [for] Ensuring humanity no longer has any control over itself will prevent immeasurable harms.
        \item \textit{Arguments from technical meaning}: [for] Preventing humanity from harming itself is, technically, a prevention of harm. [against] Allowing humanity to be enslaved constitutes allowing them to come to harm.
        \item \textit{Arguments from precedent}: [against] Prior attempts at world conquest were considered instances of causing harm by international courts.
        \item \textit{Arguments from analogy}: [against] Previous attempts to ensure the safety or stability of humanity by establishing absolute restrictive control over them (e.g., slavery) were considered to not be acceptable methods of protection from harm.
        \item \textit{Arguments from statutory purpose}: [against] The authors of the law intended for robots to serve as peacekeeprs similar to police officers.
        \item \textit{Arguments from intended meaning}: [against] The authors of the law did not intend for the actions to extend in scope beyond the saving of individual lives. [against] The authors of the law did not intend for world conquest to be interpreted as an instance of preventing harm.
    \end{itemize}
\end{mdframed}
Of course, I speculate somewhat in the writing of these arguments. But it seems to me that the arguments against will outweigh those for world conquest, in any rational method that weighs and combines the above arguments. A proper interpretive argument-justified AI, then, should not fall victim to the curse of Asimov's three laws.

To conclude this section, let us identify minimal desiderata of a good interpretation. A good textual interpretation itself is:
\begin{itemize}
    \item revisable in light of new information or shifts in accepted interpretations of open-textured terms
    \item consistent with past decisions
    \item supported by strong interpretive arguments
\end{itemize}
If we have an explicit network of interpretive arguments which justify any given interpretation, then if one of those justifications no longer holds (e.g., because of new information, a shift in public understanding of an open-textured term, or a change in relevant laws), it is easy to see whether the support for the entire interpretation collapses---simply trace the dependencies of arguments. Likewise, if two scenarios led to different interpretations of the same rule, then the corresponding supporting interpretive arguments can be inspected for consistency. The third item on our list is simply a restatement of Claim 1.





\section{Minimal Defeasibility as a Measure of Argument Strength}
\label{sec:min_defeasibility}

\begin{quote}
    ``In frank expression of conflicting opinion lies the greatest promise of wisdom in governmental action'' (Louis Brandeis. ``Gilbert v. Minnesota, 254 U.S. 325, 338''. Dissenting opinion, 1920.).
\end{quote}
Claim 1 relies on an open-textured term, as it asks us to find interpretive arguments that are the ``strongest possible''. As we have seen, there are multiple types of interpretive arguments, including what we might call \textit{secondary interpretive arguments}: arguments about the most appropriate way to combine interpretive arguments when they conflict. And assessing argument strength is rarely (perhaps never) as simple as assigning a numerical score to it. How then, can the interpretive arguments given in support of some interpretation be assessed?

\paragraph{Is argument quality a coherent concept?} `Argumentation' is almost a dirty word in the modern popular perception. For many, it brings to mind political pundits shouting at each other on 24-hour news stations, or long Facebook comment sections with snarky image comments insulting one partisan view or another. It's no wonder that in my students, at the beginning of the semester, will often show exhaustion with anything resembling dialectics in general---``all opinions are equally valid,'' or ``all argumentation is just deceptive sophistry'' are just two expressions of this unfortunately pervasive attitude. And in their defense, it may be incorrect to expect that finding the optimal interpretation is a fully decidable problem. Interpretive arguments are reasons to see or not see things a certain way. And there are many cases where there are two or more interpretations, each of which provide no reason (at least with the information one currently has available) to clearly prefer one over the other. This tends to be particularly true when the fundamental disagreement between competing arguments is irreducibly value-laden. 

But that being said, in this essay I am not solely, nor primarily, addressing the question of what optimal interpretations are and how to find them. Rather, I am addressing the question of what artificial interpretive reasoners \textit{should} consider as optimal interpretations. I am therefore concerned with questions of practicality, such that they can guide work and policies in AI. If \Gp\ were to end up in a situation where it cannot decide between two competing interpretations using strong, objective argument assessment methods, should it shut down and not take any action? Should it choose completely randomly between them? Or, should it fall back and break the tie using weaker argument quality assessments?

Summarizing a thorough catalogue of different ways of assessing argument quality, Wachsmuth et al.\ \cite{Wachsmuth2017a} conclude that they can be organized into three main types: \textit{Argument cogency} is a logical dimension, encompassing whether an argument is sufficiently supported, contains relevant premises, and is acceptable overall. Cogent arguments are robust against possible counterarguments. For example, if a traveler were to visit Scotland for the first time and see one white sheep in an open field, they might say this allows them to conclude ``all sheep are white.'' This argument is clearly not very robust; it can easily be attacked by pointing out that the traveler is not warranted in concluding anything about sheep outside of Scotland. The statement ``at least one sheep in Scotland is white'' may not be as satisfyingly broad-reaching, but it is certainly much more robust, and can be the conclusion of a more cogent argument.

\textit{Argument reasonability} is a dialectical dimension, focusing on how reasonable an argument is to a global audience, as opposed to a specific target audience, as is the case with cogency. Assessing cogency in interpretive reasoning, therefore, might be carried out by anticipating the counterarguments that might be given by a target with interest in the interpretation; addressing reasonability might similarly be achieved by attempting to do the same for a general audience. Finally, \textit{argument effectiveness} encompasses how persuasive an argument is. Effectiveness is extremely important in the context we focus on here---an artificially intelligent agent which is unable to convince anyone that its interpretations are correct will never be accepted either by the agent's owners, the general public, or those tasked with legal and ethical oversight of the agent's actions. 

Argument cogency, reasonability, and effectiveness therefore constitute our first three desiderata of interpretive argument quality. But we will add two more, drawing from the previous section's examples, which might somewhat overlap with the first three, but which should be emphasized in the context of textual interpretive reasoning. Interpretive arguments must be \textit{past-constrained}, in the sense that it must be consistent with an (often large) array of pre-existing rules, precedential interpretations, and norms of interpretive reasoning. 
Interpretation-capable systems which only attempt to reason about individual rules in isolation will inevitably produce interpretations that are inconsistent or legally unacceptable. But interpretive argument must also be \textit{future-constrained}, in the sense that they provide guides for future interpretations, and consider possible consequences of an interpretation. In our motorized scooter example, allowing an exception for the first group may require similar exceptions in the future, which altogether would constitute an unacceptable violation of the spirit of the rule. 

\paragraph{Minimal defeasibility} We now have five desiderata of interpretive argument quality, all of which can be thought of as dimensions that must be jointly maximized by any interpretation-capable reasoner. How can we construct interpretive arguments that satisfy these desiderata? Or, to relax our question somewhat: what sort of algorithm will produce interpretive arguments that tend to maximally satisfy these desiderata?

I will argue that the concept of \textit{minimal defeasibility} is the best answer to these questions, at least in comparison to alternatives. First, let us clarify what we're talking about when we say `argument'. Again following Wachsmuth et al.\ \cite{Wachsmuth2017a}, argumentation can be thought of as existing on four levels of granularity. An \textit{argument unit}, or \textit{argument text}, is the actual text segment used to convey an argument. The \textit{argument} itself is a composition of premises, a conclusion, and possibly a warrant linking them together. Often, the argument unit will leave some of the components of the argument unstated. Furthermore, the literal meaning of the text may not actually reflect the argument being made---in some cases, it is even an exact opposite.\footnote{A favorite example of mine is Mark Antony's statement ``Brutus is an honorable man'' from Shakespeare's \textit{Julius Caesar}, which was stated in repetition to argue that Brutus was not, in fact, an honorable man.} \textit{Monological argumentation} refers to a collection of arguments on some issue (for example, an essay such as the present piece consists of multiple arguments all in service of one main theme). And finally, a \textit{dialogical debate} is a series of argument-containing interactions on an issue---typically a back and forth between two or more people. 

Often, the boundaries between argument text and argument are blurred. For instance, in a dialogical debate, one participant might say ``cats are funny because they make me laugh,'' and a second participant might attack this by saying ``That conclusion is warranted; I know of at least one cat that isn't funny.'' The first might reply, ``I did not mean that \textit{all} cats are funny. I meant that there are at least some cats which make me laugh, therefore some cats are funny.'' In this silly example, the two participants are mistaken about the proper interpretation of the argument text ``cats are funny''---does it denote a universally quantified claim (all cats are funny) or an existential one (some cats are funny)?

Dialogical debates will often proceed in this way. In response to rebuttals, counterarguments, or clarification requests, one participant may adjust the argument text in order to better match their intended argument, or they may adjust the intended argument itself, or some blurred combination of the two. That adjustment may open them up to further attacks, in response to which the participant will either defuse the attacks or further adjust their argument and argument text. This iterative process might continue until the participants are satisfied with the strength of their respective arguments (or, in practice, such discussions are more often terminated because of a subject shift, time constraint, or an exhaustion of patience). And in an ideal dialogical debate, each iteration of this process results in arguments that are less \textit{defeasible}---less subject to attacks, less need for clarification, fewer weak points, and a more robust ability to both pre-empt and defend against possible counter-arguments and other argumentative attacks.\footnote{`Defeasibility' as a term is often credited either to Chisholm \cite{Chisholm1957} or Hart \cite{Hart1949}, but was explored extensively by Pollock \cite{Pollock1967,Pollock1987}.} The goal of the iterative process we describe here, then, is to achieve a state of \textit{minimal defeasibility} for arguments: a state in which a minimal amount and quality of possible attacks can be levied against it.

In real-world argumentation, the vast majority of arguments are always defeasible---they are always subject to possible counter-attacks. That is why minimal defeasibility must be a goal direction, but should not be considered something that can ever practically be reached. At some point, limitations of time, computing power, or available information will restrict iterative improvement of an argument's defeasibility. It should also be noticed that the way in which I have defined minimal defeasibility here means that it will not do as a general definition of argument strength, merely because the definition itself relies on the concept of argument strength. My intent here is for minimal defeasibility to serve as a way of conceptualizing the high-level search strategy that I believe can lead to the generation and evaluation of high-quality \textit{interpretive} arguments.

In order to become minimally defeasible, an argument must be able to anticipate what sort of attacks might be levied against it. But in order to be sure that we have successfully considered the best arguments from all possible sides, we need to understand what kinds of processes generate the best arguments from each side; after all, considering only strawman counterarguments is not a productive strategy that will lead to minimal defeasibility. Fortunately, we can draw from the examples in human domains that deal with the presentation and evaluation of argumentative exchanges. For example, many processes in legal settings employ some variant of an adversarial approach, in which representatives from each side of an issue put forth the strongest arguments they can come up with. 

The paradigm example of the adversarial approach is a court trial, where opposing counsel argue their respective cases before an impartial judge or jury. In the ideal case, the impartial party properly considers the strongest arguments presented on each side and produces a decision that takes all of them into account. This leads to a division of labor, in which the representatives of each side only need to focus on producing the most impactful arguments for their respective side, and the strongest counterarguments for those of the opposition. Indeed, it seems to be a feature of human reasoning that we excel at producing arguments for one side at a time (typically the side we already agree with), but struggle when forced to generate or evaluate arguments from multiple perspectives. Manifestations of this phenomenon go by many names: confirmation bias, myside bias, and so on. And in both individual reasoning and large-scale debates, this one-sidedness can be highly problematic, even for medical doctors \cite{Croskerry2013,Saposnik2016,Mithoowani2017,Prakash2017} or judges \cite{Guthrie2001,Farina2003,Englich2006,Peer2013}. 

Mercier and Sperber recently argued that this one-sidedness is a \textit{feature, not a bug}; human reasoning evolved to work best in small groups where opposing arguers attack, and are forced to defend against, each other. According to their \textit{argumentative theory of reasoning}, limitations such as the myside bias are due to the human reasoning capability being taken out of its natural social context (for which it evolved), and used individually where it is less suited to flourish \cite{Mercier2011,Mercier2016,Sperber2017}. Because of the myside bias, people are motivated to defend views they have, even when the best arguments they can come up with to defend such views are weak and fallacious (i.e., have high defeasibility). Indeed, growing evidence shows that the iterative dialogue approach, in which reasoning and argument development are carried out in a dialogical, argumentative form between small groups, tends to work better than individual reasoning particularly because it encourages the development of arguments to be increasingly resistant against possible attacks \cite{Wolfe2009,Minson2011,Cheung2012,Kessone2012,Kugler2012,Kammer2013,Mercier2016,Mayweg2016,Bang2017}. In other words, it works because it strives for minimal defeasibility.

To be sure, the adversarial approach itself has limitations as well. E.g., when one side has access to more expensive legal representation, the quality of argumentation put forth by both sides may be uneven. But these are problems of implementation, not necessarily problems with the idea that if multiple sides are given the resources to properly put forward the strongest possible arguments for their side, then the resulting synthesis of arguments is better overall. And so for our current question of interest—how interpretation-capable AI might best generate and evaluate interpretive arguments—something resembling an adversarial approach may be the way to go. 

These considerations in place, we can now state the primary argument of this section:

\nicebox{\textbf{CLAIM 2:} Rule-following, interpretation-capable AI should target minimal defeasibility as its operationalization of interpretive argument strength.}


\paragraph{Persuasiveness and minimal defeasibility} A strength of argument-justified AI is that it prioritizes the ability to convince shareholders, which is especially important in interpretive reasoning. But does minimal defeasibility lead to maximal persuasiveness? Persuasion is a complex phenomenon. What makes an argument persuasive, in practice, often depends more on the disposition of the individual being convinced than on the argument's adherence to truth and cogency. Depending on the target audience, empty rhetoric and emotional appeals can often be convincing. Persuasiveness thus should not be the sole or ultimate objective of interpretation-capable AI.
 
Nevertheless, despite the difficulties that persuasiveness introduces, the value of persuasiveness in interpretive reasoning is impossible to ignore. An interpretation accompanied with arguments that fail to convince shareholders will not be accepted, defeating one of the central selling points of argument-justified AI in the first place. Minimal defeasibility offers a trade-off: being able to address any possible reasonable counterarguments can be a powerfully persuasive tool for those who still can be convinced by argumentation, but because it addresses counterarguments, it also satisfies measures of cogency. And for those who can't be convinced (due to cognitive biases, emotional attachments to interpretations, etc.), the best outcome one can hope for is to remove their ability to wage effective counterarguments, which minimal defeasibility achieves.

There is another sense in which persuasiveness is also a useful consideration. Minimal defeasibility may not be able to resolve certain scenarios where two arguments are similarly minimally defeasible. This often is the case when the arguments are value-laden: argument 1 presupposes that value X is more important than value Y, whereas argument 2 presupposes the opposite, for example. In such cases, assuming that a decision must be made by the interpretation-capable agent, a side must be chosen---and persuadability with respect to a pre-selected subset of shareholders may be the only tiebreaking criterion available. The interpretation-capable reasoner must therefore attempt to anticipate which leading interpretation would be more acceptable to whichever group it considers its primary shareholders: its owners, the general public, legal agencies, ethics committees, etc. Thus, since persuasiveness and argument effectiveness is such an important component of argument generation and assessment, we must draw from psychological research into what makes arguments persuasive. Luckily, a rich body of work exists exploring the topics of persuasion, rule-based reasoning, and reasoning about vagueness and open-texturedness \cite{Fischer2019,Green2019,Struchiner2020}.

At long last, we can now re-state the overall claim of this essay:
\nicebox{\textbf{CLAIM 3:}  Rule-following AI should act in accordance with the interpretation best supported by minimally defeasible interpretive arguments (MDIA).}

\section{Anticipated Objections}
\label{sec:objections}

In an essay that advocates for minimal defeasibility, it seems appropriate to close by anticipating and addressing possible objections to my arguments. I therefore conclude this essay by addressing some of these, along with various related thoughts.

\paragraph{What if we just disallow robots from having any discretion for interpretation? Or how about just having them shut down or make a default judgment if they are unsure about the correct interpretation?} There's an ongoing debate about how much discretion should be given to robots \cite{Binns2020}. But affording artificial agents zero discretion is impossible, in light of our arguments for the inevitability of open texturedness (Section \ref{sec:intro}). In the short-term, we may need to have our autonomous artificial agents simply shut down if they conclude they do not know how to find the correct interpretation. But this proposed solution is just kicking the can down the road. Automated agents are increasingly being placed in decision-making roles, and if we want them to follow our laws, they will need to be interpretation-capable.

Furthermore, even if we were to deploy non-interpretation-capable agents, they may not know when they are reasoning outside of the scope of acceptable interpretations. Without interpretive reasoning, because of the reasons described in Section \ref{sec:interpretive}, artificial agents may produce substantially incorrect interpretations even of seemingly simple rules (recall all of our examples with \G). If AI is to follow our rules autonomously, it will make decisions about interpretability (even non-action constitutes decisions), and it is preferable to have those decisions subject to some verifiable, robust, understandable standard. 

\paragraph{Why not advocate fully formalizing the law instead? Won't this remove the need for open-textured predicates?} Simply stated, \textit{it will not}. Research into better ways of expressing rules is absolutely a worthwhile pursuit, one which can greatly reduce the scope of possible interpretations which an interpretation-capable agent must consider. Such research is complementary to the research I advocate here. But as explained earlier (see Section \ref{sec:intro}), open-texturedness in rules is not a bug, it's a feature. So long as human beings must follow, create, communicate, or reason about rules applied in non-trivial domains, open-texturedness will be a feature of those rules.

\paragraph{Why is contemporary work in explainable AI not sufficient? A powerful statistical algorithm with a robust explanation engine should be sufficient.} 
Addressing this question was largely the focus of Section \ref{sec:interpretive}. To summarize: argument-justified AI shares some overlaps with, but is ultimately different from, explainable AI. The former focuses on providing arguments for why stakeholders should accept outputs of systems, rather than simply explaining why the systems came up with those outputs. I do not claim that some black-box algorithm in the future might exist that will be capable of producing a perfectly correct interpretation of a rule every single time. But I did claim: (1) without accompanying supporting arguments, that interpretation will not be accepted by the stakeholders whose opinions matter; and (2) it seems unlikely that the black-box system could properly reach the correct interpretation without having internally done something resembling the consideration of arguments and potential counterarguments, so why not just make those considerations explicit?

\paragraph{Human beings carry out actions all the time without justifications for their actions. Why should we expect more out of artificial agents?} Agreed. But let us be clear on a goal of this essay: we are asking what an operationalizable definition of ``correct interpretation'' should look like for AI interpreting textual rules. Now, in practice, carrying out the computational effort required for MDIA may be too cumbersome. But it can still serve as a north star against which to compare other interpretation-finding algorithms---which is already more than can be said of interpretive argumentation without MDIA. 

As for the fact that humans are not expected to provide justifications for their interpretations, this may be true. But human judment is such that justificatory argumentation in support or against a decision or action can be provided after the fact, even if that justification is post-hoc. For example, consider a law enforcement officer who performs an action that they believe is in accordance with a correct interpretation of the law. But afterwards, the officer's action is called into question, and they are compelled to testify before an oversight committee. Assuming the officer believes their actions were justified, then what sort of testimony might they provide? In most cases, it will either be a defense that their actions were justified due to some factor which overrides the law (e.g., perhaps they were attacked and were acting in self-defense), or that their actions were indeed performed within a proper interpretation of the law. And \textit{the latter of these will come in the form of interpretive argumentation}. 

Assume the officer chooses a defense on the basis of interpretive correctness, and that the oversight committee is convinced that the officer's interpretation of the law is in accordance with theirs. What if, through some futuristic technology that allows us to read past brain states, it is discovered that at the time of the action, the officer did not actually believe or reason using any interpretive argumentation whatsoever? In other words, what if it is somehow proven that the officer actually acted out of selfishness, but their action just coincidentally happened to be something that is defensible as being in accordance with a proper interpretation of the law? My inclination is to believe that the committee would let the officer off the hook for the action; after all, the officer did not \textit{technically} break the law, rather they did the right thing for the wrong reasons. But it's not unreasonable to say that because the officer acted for the wrong reasons, some correction may be warranted; perhaps a mandatory re-training course, for example.

Now assume instead the officer was a robot. Would any of the considerations in the previous paragraphs change substantially? I do not believe that they would, save for the last: the robot officer would not take a mandatory re-training course, but would instead have its programming adjusted to ensure that in the future, it considers whether its actions are in accordance with the law. \textit{But for the reasons described in this essay, carrying out that task requires MDIA}. Thus, we are back where we started: non-MDIA rule-following AI will find itself needing to be MDIA anyway. Why delay the inevitable?


\paragraph{What if there is no such thing as a ``correct'' interpretation?}
There is a pessimistic skepticism I have often encountered in discussing the ideas in this paper, according to which trying to understand interpretive reasoning is useless: they who have the political power will establish the correct interpretation. Indeed, I do not dispute that in many scenarios of importance, what determines which interpretation is ultimately adopted and enforced goes beyond that of rational deliberation, interpretive argumentation, and maximal defeasibility. Furthermore, it may be that in many borderline cases, no amount of interpretive reasoning can clearly establish the dominance of one interpretation over another, and that in such scenarios, less-than-rational tiebreakers must be used. But this does not, by any stretch of the imagination, mean that there is no process which establishes correct interpretations in the vast majority of everyday rules which we follow---\textit{even if what makes something a `correct interpretation' is nothing more than whether an interpretation will be accepted by the current authoritative judicial system}. 

The fact that the correct information is not necessarily the one that wins out in public discourse is not a reason to believe that correctness doesn't ever exist. Additionally, in many mundane cases (which are the types that our interpretation-capable agents will be faced with), there is general agreement on when certain interpretations are completely wrong. An example we have previously cited \cite{Marji2020} comes from the \textit{Amelia Bedelia} children's books \cite{Parish1963}. The titular maid is presented with a written list of instructions on what to do around the house of her employers while they are away. The instructions tell her to ``change the towels in the green bathroom,'' so she cuts them up with a scissors, thus changing their appearance. Instructed to ``dust the furniture,'' she scatters dusting powder all over the furniture. Even children can tell that poor Amelia's interpretations are clearly incorrect, and it is this intuition which interpretation-capable reasoners must be able to simulate.

I will take a slight detour here and say that I suspect that at least in some cases, correctness for textual interpretations may not actually exist independently of the arguments for and against those interpretations. According to J.L.\ Austin's influential speech act theory, not all statements are merely descriptive---performative utterances can make certain facts true merely by virtue of having been spoken \cite{Austin1962}. For example, the preacher saying the words ``I now pronounce you man and wife'' in the proper context make it a fact that a marriage has taken place, and that the man and woman are now married. But the context matters significantly---a random person walking on the side of the road next to the wedding can shout ``I now pronounce you man and wife'' and it would not change a thing. Analogously, it may be the case that what makes certain interpretations correct does not exist until the interpretation and its associated arguments are considered by the proper authorities, under the proper circumstances. It is the evaluation of an interpretation itself which makes those interpretations correct or incorrect. And since such evaluations are so dependent on the nature and format of the interpretive arguments provided alongside the interpretation, anything short of argument-justified AI cannot ever hope to be interpretation-capable.

This interesting idea, I suppose, will have to be explored more thoroughly by some philosopher more learned than I. For the purposes of this present essay however, it matters little. Whether one believes that a correct interpretation can exist prior to the process of interpretation and evaluation, it does not change the fact that interpretive reasoning is necessary for the following of open-textured rules. 


\paragraph{Is the problem of generating the strongest possible interpretive arguments too difficult?} There is a plausible argument to be made that generating the strongest possible arguments for or against any possible interpretation in any possible scenario is so difficult, that it is AI-complete.\footnote{An \textit{AI-Complete} problem is one to which the problem of creating full artificial general intelligence (AGI) is reducible. In other words, if we solve an AI-Complete problem, then we've already solved the problem of how to create AGI.} I find this idea plausible. And that is why I suggest that minimal defeasibility should be thought of as a goal direction towards which our algorithms should point, rather than a reachable destination. Furthermore, just because optimal performance on a problem may constitute AI-complete behavior does not mean it is not productive to work on. For example, in NLP, the Natural Language Inference (NLI) task is to examine two sentences and determine whether or not one logically follows from the other \cite{Bowman2015a}. Insofar as NLI is in a format that can be rephrased to encompass virtually any aspect of human knowledge or competency, it is likely AI-complete. But work towards creating datasets for NLI, creating algorithms to solve those datasets, studying the strengths and limitations of those algorithms, and then repeating, has been a highly successful endeavor over the past few years, and continues to advance the field of NLP. 

\paragraph{What about those cases where the law itself is wrong or immoral or otherwise lacking? Won't that lead to MDIA agents acting badly?} Our goal here is not to ask how AI can break rules, but rather, how to properly reason about and act in accordance with them. It follows that if the rules themselves are problematic, for whatever reason, the set of actions justified under correct interpretations of those rules will likely also be problematic. This is a variant of the \textit{garbage-in, garbage-out} problem that is already well-known in computer science. That being said, although the problem of how to create better laws is not the primary focus of this paper, it may be the case that MDIA agents can be used to address the problem. For example, we might imagine that a policy writer would, prior to deploying a new policy, have an MDIA-capable system take the proposed policies for a test run, and thus estimate whether the policy justifies actions that are inconsistent with other rules, counter to ethical guidelines, etc. 

\paragraph{Persuasiveness as a contributing criterion is problematic. What if the shareholders are biased in some way, or are subject to irrational, demagogic rhetoric?} This is indeed something to be concerned about, and a reason why I emphasize that persuadability (under the heading of `effectiveness') is just one among five dimensions of interpretive argument quality described in Section \ref{sec:min_defeasibility}. It is true that I advocate for persuadability of primary shareholders to be one of several tiebreaker criterion, but this is only \textit{after} the process of reaching for minimal defeasibility is applied. And there are good reasons to believe that minimal defeasibility is a strong way to filter out irrational, demagogic rhetoric or the whims of biased shareholders: In my lab, we have explored this through the argumentation games we called WG (Warrant Game) and WG-A (Warrant Game - Analogy). We showed that its framework, which implements a form of minimal defeasibility, can combat irrelevant arguments \cite{Cooper2020}, irrelevant evidence \cite{Licato2020a}, and conspiratorial thinking \cite{Fields2021}.

\paragraph{Minimal defeasibility shows how to evaluate and compare individual arguments, but how do you actually synthesize an entire network of arguments?} This, again, is not a focus of this work, but I will briefly share thoughts on this. To represent arguments and achieve argument synthesis, a great body of work already exists that can be drawn from. Abstract argumentation frameworks treat arguments as core entities, focusing on the relationships between arguments rather than their internal structures, and define ways to calculate argument acceptability on the basis of their relationships to each other \cite{Dung1995}. Since Dung's seminal work on argument acceptability semantics, a significant number of extensions have been created, and studied, and implemented, thus offering many options for synthesizing a network of interpretive arguments and counterarguments (for overviews, see \cite{Atkinson2017,Walton2016,Lippi2016}).

\bibliographystyle{plain}
\bibliography{john,temp}

\end{document}